\documentclass[lettersize,journal]{IEEEtran}
\usepackage{amsmath,amsfonts}
\usepackage[ruled,vlined]{algorithm2e}
\usepackage{array}
\usepackage[most]{tcolorbox}
\usepackage{textcomp}
\usepackage{siunitx}
\usepackage{stfloats}
\usepackage{pgfplots}
\usepackage{pifont}
\pgfplotsset{compat=1.17}
\usepackage{xcolor}
\usepackage{helvet}
\usepackage{url}
\usepackage{tabularx}
\usepackage{sansmath}
\usepackage{verbatim}
\usepackage{graphicx}
\usepackage{cite}
\usepackage[normalem]{ulem}
\usepackage{enumitem}
\usepackage{threeparttable}
\usepackage{makecell} 
\usepackage{multirow}
\usepackage{booktabs}
\usepackage{subcaption}
\usepackage{fontawesome} 

\newcommand{\cmark}{\textcolor{green!70!black}{\checkmark}} 
\newcommand{\xmark}{\textcolor{red}{\ding{55}}}             

\usepackage{etoolbox}
\definecolor{mypurple}{RGB}{139, 68, 227} 
\definecolor{myyellow}{RGB}{250, 190, 23} 
\definecolor{myorange}{RGB}{255, 81, 1} %
\definecolor{mygreen}{RGB}{0, 214, 105} %
\definecolor{myred}{RGB}{255, 50, 179} %
\begin{document}

\title{Reinforced Strategy Optimization for Conversational Recommender Systems via Network-of-Experts}


\author{
Xiaoyan~Zhao$^\dag$,
Ming~Yan$^\dag$,
Yang~Zhang,
Yang~Deng,
Jian~Wang, \\
Fengbin~Zhu,
Yilun~Qiu,
Hong~Cheng,
and~Tat-Seng~Chua
\IEEEcompsocitemizethanks{
    \IEEEcompsocthanksitem X. Zhao and H. Cheng are with The Chinese University of Hong Kong, Hong Kong, China. E-mail: \{xzhao, hcheng\}@se.cuhk.edu.hk.
    \IEEEcompsocthanksitem M. Yan is with the University of Science and Technology of China, Hefei, China. E-mail: ym689@mail.ustc.edu.cn.
    \IEEEcompsocthanksitem Y. Zhang, F. Zhu, Y. Qiu, and T. Chua are with the National University of Singapore, Singapore. E-mail: \{zhangy, fengbin, qiuyilun, dcscts\}@nus.edu.sg. 
    \IEEEcompsocthanksitem Y. Deng is with the Singapore Management University, Singapore. E-mail: dengyang@smu.edu.sg.
    \IEEEcompsocthanksitem J. Wang is with The Hong Kong Polytechnic University, Hong Kong, China. E-mail: jian.wang@polyu.edu.hk.
}
\thanks{$^\dag$~X. Zhao and M. Yan are co-first authors.}
\thanks{Manuscript received XXXXXX XX, XXXX; revised XXXXXX XX, XXXX.}
}

\markboth{IEEE Transactions on Knowledge and Data Engineering,~Vol.~X, No.~Y, Month~2024}%
{Your Name: Mixture-of-Experts for Conversational Recommender Systems}


\maketitle

\begin{abstract}
Conversational Recommender Systems (CRSs) aim to provide personalized recommendations through multi-turn natural language interactions with users. Given the strong interaction and reasoning skills of Large Language Models (LLMs), leveraging LLMs for CRSs has recently emerged as a promising direction.
However, existing LLM-based methods often lack explicit optimization of interaction strategies, instead relying on unified prompts and the LLM’s internal knowledge to decide how to interact, which can lead to suboptimal outcomes. In this paper, we propose a novel \underline{\textbf{R}}einforced \underline{\textbf{S}}trategy \underline{\textbf{O}}ptimization (\textbf{RSO}) method for CRS, which decomposes the process of generating strategy-driven response decisions into the macro-level strategy planning and micro-level strategy adaptation through a network-of-experts architecture. At the macro level, a \textit{Planner} expert selects macro-level interaction strategies (e.g., recommend, explain, encourage). 
At the micro level, an \textit{Actor} expert generates detailed responses conditioned on the selected macro-level strategy, guided by auxiliary experts that provide complementary information such as user preferences and factual grounding. This hierarchical decomposition disentangles the optimization of different sub-tasks involved in CRS response generation, enabling more tractable learning at each level. To address the scarcity of high-quality multi-turn training data, we formulate strategy learning as a reinforcement learning problem, guided by an LLM-based reward model to achieve automatic strategy exploration. Extensive experiments show that \textit{RSO} significantly improves interaction performance compared to state-of-the-art baselines, demonstrating the effectiveness of explicit hierarchical strategy optimization for CRS. 
\end{abstract}


\section{Introduction}
\label{sec:intro}
\begin{figure}[h] 
    \centering
    \includegraphics[width=\columnwidth]{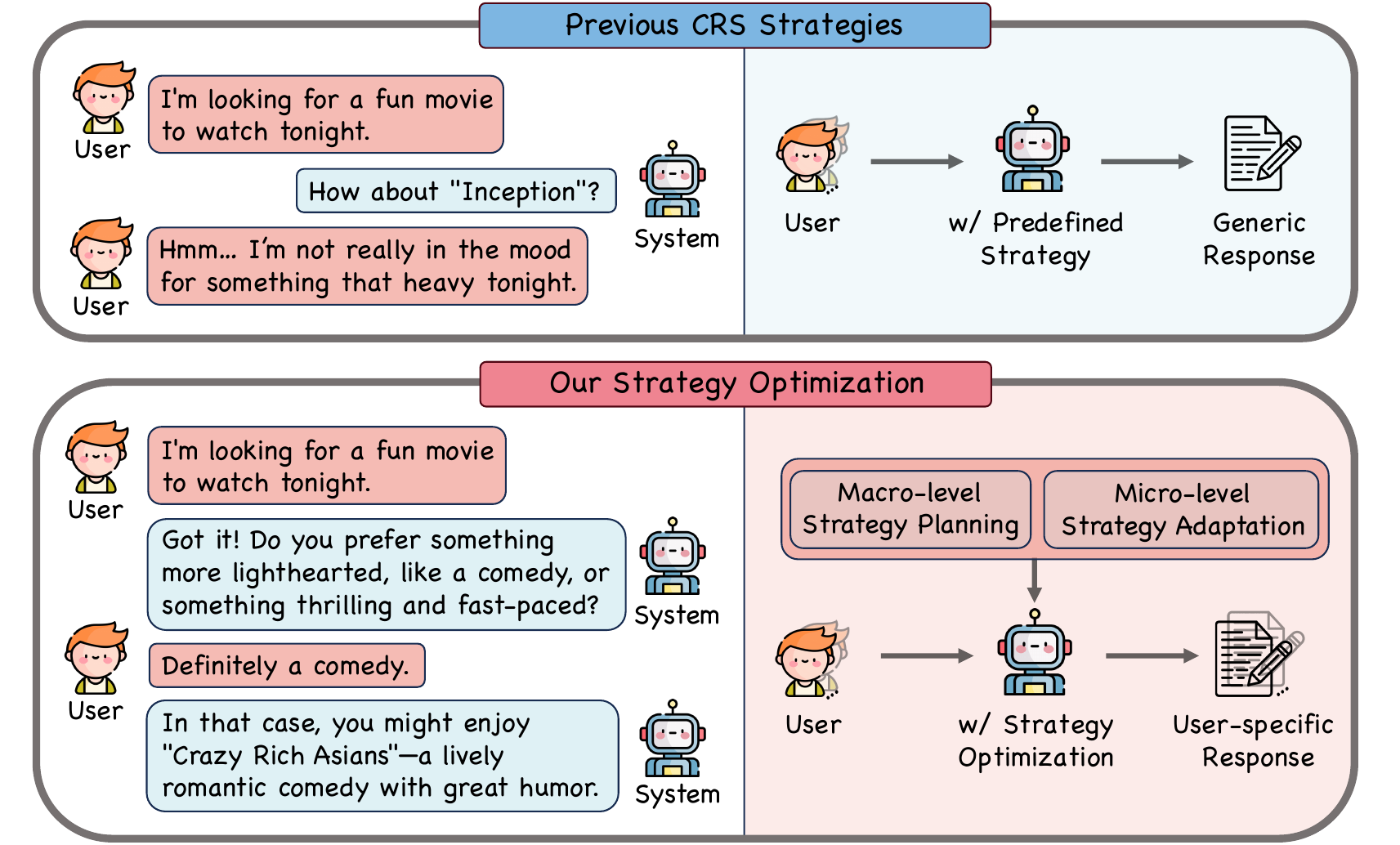}
    \caption{Comparison between previous CRS strategies and our proposed strategy optimization. The right panel illustrates the difference in a single system response: previous methods rely on prompt-based predefined strategies~\cite{pccrs,inspired} and yield generic responses, while our approach performs explicit strategy optimization to generate user-specific responses.}
    \label{fig:intro_img}
\end{figure}

Conversational Recommender Systems (CRSs)~\cite{lei2020conversational,CRSsurvey,deng2021unified,gao2021advances,DBLP:journals/tkde/WangJGZW24,DBLP:journals/tkde/DaiWXYL24,DBLP:journals/tkde/YuLWX25} aim to interact with users through natural language conversation, elicit their preferences, and refine recommendations to maximize user satisfaction and acceptance of the recommendations.
Different from traditional recommenders~\cite{recsurvey,sasrec,bert4rec,lightgcn,slmrec,binllm,hu2025exact,collm} that rely solely on static historical data,
CRSs need to determine not only what to recommend but also how to strategically interact with users throughout the conversation—deciding when to ask, how to elicit
preference, and how to justify responses.
These strategy-driven response decisions critically influence the effectiveness and efficiency of the interaction, ultimately shaping the overall user experience~\cite{yu2011toward,alslaity2019towards,zhang2020evaluating}.

Given the strong interaction and reasoning capabilities of Large Language Models (LLMs)~\cite{gpt4,qwen25,llama3}, leveraging LLMs for CRSs has become a promising direction, with several notable efforts emerging~\cite{feng2023large,wang2023rethinking,deng2024towards,xi2024memocrs,zhu2025collaborative}.
However, as shown in Fig.~\ref{fig:intro_img}, existing approaches~\cite{pccrs,inspired,li-etal-2025-chatcrs} often overlook explicit strategy optimization, which is essential in CRS for dynamically deciding how to steer multi-turn conversations toward engaging interactions.
Without such optimization, systems typically rely on fixed prompts and the LLM’s internal knowledge to decide interaction behavior, which often yields less user-adaptive behaviors, e.g., recommending too early, failing to elicit enough preference information, or providing ungrounded information.
These limitations lead to suboptimal strategy planning, resulting in inefficient conversation, failure to capture user intent, and ultimately degrading the overall user experience.

This work considers introducing explicit strategy optimization for LLM-based CRS, enabling more targeted and adaptive interaction strategies for different users. However, this is a non-trivial task. 
First, each interaction decision must jointly account for critical determinants of user satisfaction, including preference reasoning, factual grounding, and dialogue context, which together create a large and entangled decision space. 
Directly optimizing strategy selection within this space is challenging, particularly as current LLMs lack explicit mechanisms for personalized interaction planning.
Second, the multi-turn nature of CRS further complicates learning. It is challenging to obtain high-quality supervision signals for optimal strategies across entire dialogue trajectories. As a result, pure supervised learning is often insufficient.

To address these challenges, we propose a hierarchical decomposition approach to strategy optimization in CRS. 
The process of generating strategy-driven responses is decomposed into two complementary levels: \textit{macro-level strategy planning} and \textit{micro-level strategy adaptation} to user-specific contexts.
At the macro level, the system determines the overall conversational strategy, selecting which type of action to take (e.g., recommend, explain, encourage) based on the conversation state and user feedback.
At the micro level, the system operationalizes the chosen strategy by adapting it to the user’s context. This involves inferring user preferences and grounding responses in factual knowledge to deliver coherent, personalized outputs.
This hierarchical decomposition disentangles the task into two relatively independent stages, each operating within well-defined and tractable decision scopes. By learning within these disentangled objectives, we mitigate the complexity of the original entangled optimization problem and enable more robust and personalized strategy learning.

To this end, we propose a novel \underline{\textbf{R}}einforced \underline{\textbf{S}}trategy \underline{\textbf{O}}ptimization (\textbf{RSO}) method for CRS based on hierarchical decomposition. At its core, RSO introduces a hierarchical network-of-experts architecture that performs macro-level strategy planning and micro-level strategy adaptation. At the macro layer, a \textit{Planner} expert is introduced to select the macro-level strategy at each turn. 
At the micro level, an \textit{Actor} expert collaborates with multiple \textit{auxiliary experts} to operationalize the selected macro-level strategy. 
While the \textit{Planner} expert decides the type of conversational strategy, its decision alone is insufficient to produce effective user-facing responses.
User-adaptive interaction requires inferring user intent and grounding the responses in reliable information to maintain trust. Thus, we introduce two auxiliary experts: a \textit{Preference Reasoner} to infer the user’s evolving preferences, and a \textit{Fact Retriever} to ensure the factual accuracy and reliability of candidate content. Together, they provide the Actor with user-specific context, enabling coherent, grounded, and persuasive responses. 
Besides, we cast Planner training as a reinforcement learning (RL) problem rather than purely supervised fine-tuning, allowing the system to actively explore strategy space with guidance from an LLM-based reward model. This reduces reliance on scarce multi-turn annotations while enabling dynamic feedback-driven optimization. 

The main contributions of this work are summarized as follows:
\begin{itemize}[leftmargin=*]
    \item We introduce a hierarchical decomposition paradigm for the strategy optimization in CRS, leveraging the macro-level strategy planning and micro-level strategy adaptation to realize dynamic and effective CRS interaction.
    \item  We propose a reinforced strategy optimization method (\textbf{RSO}) that employs a network-of-experts to disentangle key subtasks and enable user-adaptive response within well-defined and tractable decision scopes.  
    \item We conduct extensive experiments on two representative CRS benchmark datasets, showing that our method enables more effective strategy optimization and improves both conversational quality and recommendation performance.
\end{itemize}

\section{Problem Formulation}
\label{sec:problem_formulation}
We formulate the strategy optimization in CRSs as a sequential decision-making process under the Markov Decision Process (MDP) framework. 
At each turn $t$, the CRS interacts with a user $u$ to uncover preferences and recommend relevant items through natural language dialogue. 
A session of dialogues is represented as a trajectory $\tau = \{s_1, a_1, r_1, \dots, s_T, a_T, r_T\}$, where $s_t$ is the dialog state at turn $t$, which captures the context up to current turn;
$a_t$ is the action, \textit{i.e.}, the response of the system, potentially including recommending operation or any other dialogue operation;   
$r_t \in \mathbb{R}$ is the scalar reward signal measuring the response quality towards task success. 
The goal of strategy optimization is to learn a policy model $\pi_{\theta}(a_t|s_t)$ that maximizes overall user satisfaction throughout the conversation, encompassing both recommendation quality and dialogue experience, as measured by the expected cumulative reward over the session:
\begin{equation}
    \max_{\pi_\theta} \ \mathbb{E}_{\tau \sim \pi_\theta} \left[ \sum_{t=1}^{T} r_t \right]
\end{equation}
To maximize overall user satisfaction, each response generation must jointly account for factors such as inferring current user preferences, generating dialogue content, and verifying factual information, which makes strategy optimization inherently challenging.

\begin{figure*}[t]
    \centering
    \includegraphics[width=1.0 \linewidth]{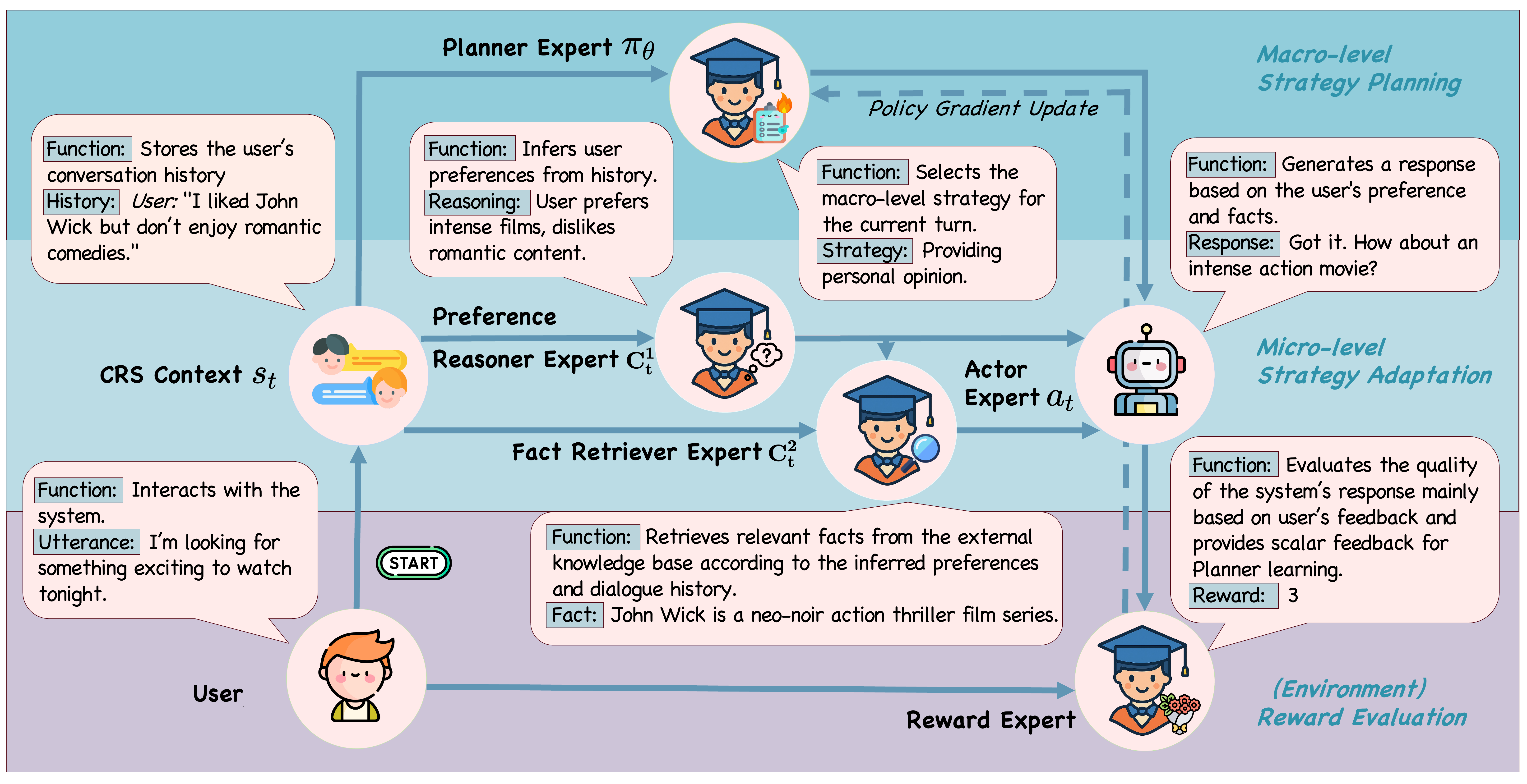}
    \vspace{-7mm}
    \caption{Overview of our RSO framework. Specifically, the \textit{Planner} performs macro-level strategy planning, while the \textit{Preference Reasoner}, \textit{Fact Retriever}, and \textit{Actor} collaboratively realize micro-level strategy adaptation. The \textit{Rewarder} provides a turn-level signal to update the \textit{Planner}.}
    \label{fig:framework}
\end{figure*}

\section{Methodology}
\label{sec:method}
We propose a hierarchical strategy optimization framework based on a Network-of-Experts (NoE) architecture to explicitly optimize interaction strategies for more personalized and adaptive CRSs. In the following section, we first provide an overview of the method, then detail the NoE-based hierarchical policy design, and finally describe the learning approach.

\subsection{Overview}

As discussed, the inherently entangled multi-factor nature of CRSs during each response generation makes direct learning $\pi_{\theta}(a_t|s_t)$ highly challenging. To simplify strategy optimization, we hierarchically decompose it into:
\begin{enumerate}[leftmargin=*]
\item \textbf{Macro-level Strategy Planning}: Determines a macro-level strategy, \textit{i.e.,} whether to recommend or to inquire about user preferences, formulated as $\pi_{\theta_{1}}(h_t|s_t)$, where $h_t$ denotes the macro-level strategy and $\theta_{1}$ is the model's parameters.
\item \textbf{Micro-level Strategy Adaptation:} Produces the final action (response) by instantiating the chosen macro-level strategy while considering additional factors such as user preferences and factual grounding, formulated as $\pi_{\theta_{2}}(a_t|h_t,s_t)$, where $\theta_{2}$ denotes the corresponding model.
\end{enumerate}

Then, the overall policy model is obtained by cascading the two levels of models:
$$
\pi_{\theta}(a_t|s_t) = \pi_{\theta_{2}}\big(a_t \mid \pi_{\theta_{1}}(h_t|s_t), s_t\big).
$$

\textbf{NoE Architecture and RL-based Learning}. 
To implement the hierarchical decomposition, we leverage a Network-of-Experts (NoE) framework illustrated in Fig.~\ref{fig:framework}, consisting of two levels of experts:
(i) a Planner Expert $\pi_{\theta_1}$ at the macro level, responsible for selecting the macro-level strategy $h_t \in \mathcal{H}$ at each turn, i.e., implementing $\pi_{\theta_1}(h_t|s_t)$; and
(ii) a micro-level Actor Expert, supported by several Auxiliary Experts, where the Actor uses the selected macro-level strategy and additional information (e.g., user preferences) from the Auxiliary Experts to produce the final action, i.e., implementing $\pi_{\theta_2}(a_t|h_t, s_t)$.

For effective training, we frame the entire process as an RL problem and propose a two-stage RL-based learning method: 1) the first stage performs SFT to warm up the model; and 2) the second stage conducts RL training to enable better strategy exploration.
During the second-stage RL training, as shown in Fig.~\ref{fig:framework}, we introduce a \textit{Rewarder Expert} that is carefully designed to reflect the overall user experience, providing turn-level scalar reward signals for model updates.

Next, we present the detailed hierarchical design within the NoE framework, starting with macro-level planning, followed by micro-level strategy adaptation. We then elaborate on the two-stage RL-based learning method.

\subsection{Macro-Level Strategy Planning}
At each turn, we first determine a macro-level strategy, \textit{i.e.}, the type of interaction to adopt, such as asking a question, offering an explanation, or engaging in casual conversation. This decision is made by a \textit{Planner Expert}.

The Planner is implemented as a learnable and lightweight language model. At each turn $t$, it can sample an appropriate macro-level strategy from a candidate set according to its generated probabilities conditioned on the current state $s_t$. Formally,
\begin{equation}
h_t \sim \pi_{\theta_1}(h_t \mid s_t),
\end{equation}
which indicates that the Planner selects $h_t$ according to the probability distribution $\pi_{\theta_1}(h_t \mid s_t)$ generated by the Planner parameterized with $\theta_1$. 
The Planner is also provided with a fixed candidate set $\mathcal{H}$ of potential strategies. 
Although omitted in the above formulation for simplicity, the policy can be equivalently expressed as $\pi_{\theta_1}(h_t \mid s_t; \mathcal{H})$. The set $\mathcal{H}$ is predefined and remains the same for all users and turns.

\begin{table*}[t]
\centering
\caption{Descriptions of CRS strategies, categorized into sociable strategies, preference elicitation inquiries, and non-strategy.}
\label{tab:strategies}
\begin{tabular}{p{0.12\textwidth} p{0.2\textwidth} p{0.6\textwidth}}
\toprule
\textbf{Category} & \textbf{Strategy Type} & \textbf{Strategy Description} \\
\midrule
\multirow{10}{*}{\parbox{0.15\textwidth}{SOCIABLE\\ STRATEGIES}} & \textsc{Credibility} & Provide factual information about the item attributes to demonstrate expertise. \\
\addlinespace[0.7ex]
& \textsc{Personal Opinion} & Express subjective opinion about the item without contradicting given factual information. \\
\addlinespace[0.7ex]
& \textsc{Encouragement} & Compliment the user's taste and encourage them to try the recommended item. \\
\addlinespace[0.7ex]
& \textsc{Acknowledgment} & Use short, cheerful responses to convey excitement or appreciation. \\
\addlinespace[0.7ex]
& \textsc{Similarity} & Express similar preferences or opinions, or show agreement with the user's views. \\
\addlinespace[0.7ex]
& \textsc{Offer Help} & Offer assistance in finding suitable recommendations. \\
\addlinespace[0.7ex]
& \textsc{Personal Experience} & Share your own experience related to the recommended item. \\
\addlinespace[0.7ex]
& \textsc{Self Modeling} & Model behavior by stating your own positive reaction to the recommended item. \\
\addlinespace[0.7ex]
& \textsc{Transparency} & Be honest about the recommendation logic or confirm user preferences before recommending. \\
\midrule
\multirow{3}{*}{\parbox{0.15\textwidth}{PREFERENCE\\ ELICITATION\\ INQUIRIES}} & \textsc{Opinion Inquiry} & Ask about the user's opinion on specific item attributes. \\
\addlinespace[0.7ex]
& \textsc{Experience Inquiry} & Ask about the user's past experiences to gather more information about their preferences. \\
\addlinespace[0.7ex]
& \textsc{Rephrase Preference} & Rephrase the inferred user preferences to confirm understanding. \\
\midrule
\multirow{1}{*}{NON-STRATEGY} & \textsc{No Strategy} & Chat naturally with the user without following any specific conversational strategy. \\
\bottomrule
\end{tabular}
\end{table*}
\textit{\textbf{Candidates of Macro-Level Strategies.}}
In this work, we follow~\cite{inspired} to define the candidate set of macro-level strategies based on common communicative goals in CRS. The complete list of strategies is presented in Table~\ref{tab:strategies}. We consider three groups of macro-level strategies:
\begin{itemize}
    \item \textsc{Sociable Strategies}: Foster engagement through casual, open-ended conversation.
    \item \textsc{Preference Elicitation Inquiries}: Elicit more fine-grained user preferences via clarification or comparison questions.
    \item \textsc{Non-strategy}: Serve as default response behaviors when no specific strategy applies.
\end{itemize}

Each strategy type is clearly defined and mapped to dialogue intents in the table, which are grounded in findings from social science~\cite{krippendorff2004reliability}.
These macro-level strategies serve as macro-level priors that guide the overall conversational flow.

\subsection{Micro-Level Strategy Adaptation}
While the macro-level strategy determines the high-level type of interaction (e.g., recommend, encourage, explain), relying solely on this coarse-grained decision is insufficient for effective conversational recommendation.
In practice, user interests are dynamic and can shift across turns, meaning that simply following a pre-selected strategy without considering the evolving dialogue context may lead to suboptimal or even irrelevant responses.
To fully capture the user’s real-time intent, it is crucial to continuously reason about the user’s preferences during the conversation, ensuring that the strategy is instantiated in a way that reflects the user’s current needs and mood.

Moreover, fact grounding is essential: grounding the response with accurate, retrieved information about candidate items not only improves recommendation effectiveness but also strengthens user trust by making the system’s reasoning more transparent.
Therefore, at each turn, we adapt the macro-level strategy to the specific user and conversational context by combining (i) the inferred dynamic preferences and (ii) the retrieved factual evidence, so that the final response is both personalized and well-supported.

As shown in Fig.~\ref{fig:framework}, we employ an \textit{Actor Expert} to instantiate the macro-level strategy as a concrete natural language response, supported by two auxiliary experts that provide essential context: the \textit{Preference Reasoner Expert} and the \textit{Fact Retriever Expert}.
The outputs from these auxiliary experts are then integrated by the Actor Expert to generate the final response.
In the following sections, we elaborate on these experts and explain how their collaboration enables micro-level strategy adaptation, resulting in responses that are context-aware, dynamically personalized, and factually grounded.

\subsubsection{\textbf{Preference Reasoner}}
Incorporating user preferences is essential to ensure that the final response is well aligned with the user. In multi-turn CRSs, however, user signals are often vague and implicitly distributed across dialogue turns. It is therefore necessary to exploit the dialogue history to infer more detailed user preferences. For example, when a user states “I don’t like that,” the system should infer the underlying negative preference type by considering the preceding conversation, rather than relying solely on the surface-level utterance. In our framework, we employ a Preference Reasoner, implemented with a general-purpose LLM, to fulfill this role. At each turn, given the conversational history, the LLM is prompted to explicitly analyze user preferences with a carefully designed instruction, emphasizing preference reasoning from a global dialogue perspective. Formally, at turn $t$, the process is defined as
\begin{equation}\label{eq:preference}
    C_t^{1} = \textit{Pref\_Reasoner}(s_t),
\end{equation}
where $s_t$ denotes the current state, including the dialogue history; $C_t^{1}$ denotes the extracted preference information, represented in textual form, which serves as one contextual input for the Actor in Eq.~\eqref{eq:actor}. 

\subsubsection{\textbf{Fact Retriever}}
Solely relying on the LLM’s internal knowledge may result in unreliable or even incorrect responses due to inherent hallucination issues. For instance, the system may recommend non-existent items or incorrectly associate items with certain features, which can cause malfunction and undermine user trust. To address this, we introduce a \textit{Fact Retriever} expert to ensure the reliability of the information.

The primary role of the Fact Retriever is to retrieve entities (mainly items) relevant to the current dialogue state along with their associated factual knowledge. To achieve this, we first apply an embedding-based similarity retrieval method to identify the top-$K$ related entities based on the current conversational state $s_t$ and the inferred user preference $C_t^{1}$ from Eq.~\eqref{eq:preference}. Next, we retrieve the factual knowledge of these entities from an external knowledge graph $\mathcal{G}$. The process can be formulated as:
\begin{equation}
\begin{split}
C_{t}^{2} &= \textit{Retriever}(s_t, C_t^{1}; \mathcal{G}) \\
&= \textit{KG\_retrieve}(\textit{entity\_retrieve}(s_t, C_t^{1}); \mathcal{G}),
\end{split}
\end{equation}
where $\textit{entity\_retrieve}(\cdot)$ denotes the embedding-based entity retrieval process, and $\textit{KG\_retrieve}(\cdot)$ denotes the subsequent knowledge graph retrieval process. $C_{t}^{2}$ denotes the final retrieved information. Incorporating it into response generation ensures that the output is factually grounded, which enhances dialogue quality and reliability, helping to enhance user trust.

\subsubsection{\textbf{Actor}}
The Actor serves as the core component responsible for instantiating macro-level strategies into concrete, user-specific actions by incorporating relevant context information. In CRSs, these actions take the form of natural language responses. We implement the Actor Expert using a general-purpose LLM prompted to accomplish the task.
At each turn, the Actor takes as input the current macro-level strategy $h_t$ together with textualized context information provided by auxiliary experts, integrates them, and produces the final response for the user:

\begin{equation}\label{eq:actor}
    a_t = \textit{Actor}(h_t; C_t^{1},\dots,C_t^{K}),
\end{equation}
where $C_t^{1},\dots,C_t^{K}$ denote the key context signals. This work focuses on two primary signals: (1) $C_t^{1}$, the inferred user preference from the Preference Reasoner expert, and (2) $C_t^{2}$, the retrieved factual evidence from the Factual Seeker expert. By combining macro-level strategy with user preference reasoning and factual grounding, the Actor bridges macro-level strategy planning and user-specific adaptation, ensuring responses that are both contextually appropriate and factually accurate.

\begin{algorithm}[t]
\small
\caption{One-Turn Reinforced Strategy Optimization of Our RSO Framework}
\label{alg:crs_moe}
\KwIn{Initial dialogue state $s_1$, max turns $T_{\max}$, 
strategy set $\mathcal{H}$, KG $\mathcal{G}$, Planner $\pi_{\theta_1}$,  $\textit{Pref\_Reasoner}(\cdot)$,  $\textit{Retriever}(\cdot)$, $\textit{Actor}(\cdot)$, $\textit{Rewarder}(\cdot)$.
}
\BlankLine
\For{$t=1$ \KwTo $T_{\max}$}{
    \textcolor{gray}{\tcp{(1) Macro-level strategy planning}}
    Sample macro strategy $h_t \sim \pi_{\theta_1}(h \mid s_t; \mathcal{H})$ \;
    \textcolor{gray}{\tcp{(2) Micro-level strategy adaptation}}
    Infer user preference $C_t^{1} = \textit{Pref\_Reasoner}(s_t)$ \;
    Retrieve fact $C_t^{2} = \textit{Retriever}(s_t, C_t^{1}; \mathcal{G})$ \;
    Generate response $a_t \leftarrow \textit{Actor}(h_t; C_t^{1}, C_t^{2})$ \;
    \textcolor{gray}{\tcp{(3) State transition \& Rewarding}}
    Observe user reply and update $s_{t+1} \leftarrow s_t$ \;
    Compute reward $r_t = \textit{Rewarder}(s_{t+1}, \textit{prompt}_{\textit{rwd}})$ \;
    \If{$\text{Stop}(r_t)=\text{True}$}{
        \textbf{break}
    }
}
Update Planner according to Eq.~\ref{eq:policy_update} \;
\end{algorithm}

\subsection{Reinforced Strategy Optimization}


To enable adaptive strategy planning for different users during CRS, one might ideally optimize all experts jointly to achieve fully coordinated learning. 
However, such joint optimization would result in a highly coupled reinforcement learning process with an extremely large search space, which makes optimization unstable, training slow, and convergence difficult. 
Moreover, the difficulty of learning varies across experts: while response generation and factual retrieval can already be handled effectively by modern LLMs, the most challenging and user-specific component is the macro-level strategy planning, which decides when to recommend, when to clarify, and how to steer the conversation. 

Therefore, we focus our optimization on the \textit{Planner}, while keeping the \textit{Actor}, \textit{Preference Reasoner}, and \textit{Fact Retriever} fixed. 
Recent studies~\cite{gpt4,qwen25,llama3} demonstrate that modern LLMs already exhibit strong capabilities in reasoning, contextual integration, and leveraging external knowledge through in-context learning. 
As a result, the micro-level experts can reliably perform their corresponding tasks through in-context learning and provide stable, high-quality contextual signals without additional training. 

Importantly, this design does not ignore micro-level factors during strategy optimization. 
At each step, the Planner’s decision is conditioned on the full conversational state, which incorporates user preferences and factual evidence extracted in previous turns. 
Consequently, micro-level signals are implicitly taken into account when optimizing the macro-level policy. 
This approach significantly reduces the complexity of joint optimization, improves training stability, and allows more efficient exploration of the strategy space while still aligning strategy decisions with user-specific context.

To this end, we propose a two-stage learning framework for the Planner. In the first stage, supervised fine-tuning is used to warm up the model by learning strong planning priors from human-annotated strategies. 
In the second stage, entropy-regularized RL tuning is applied to further explore and refine the strategy space with feedback from a reward model. 
Algorithm~\ref{alg:crs_moe} illustrates the optimization procedure for a single dialogue turn, which is iteratively applied across multiple turns in a full conversation session until termination or until the maximum number of turns is reached.

\subsubsection{\textbf{Warm-up Fine-tuning}}
Direct RL training from scratch is often unstable and prone to collapse~\cite{zhang2025reinforced}, particularly given the multi-turn nature of CRS. To address this, we first apply SFT to warm up the policy model, providing a strong initialization for subsequent RL optimization.

In this stage, the Planner is trained on annotated CRS datasets, where each dialogue turn is labeled with a suitable macro strategy. 
Let $\mathcal{D}_{sft}$ denote the annotated dataset, where each sample is represented by $(s_t, h_t^{'})$, meaning $h_t^{'}$ is a good macro-level strategy given the state $s_t$. Then the Planner model is optimized by minimizing the following loss:
\begin{equation}
    \mathcal{L}_{\text{SFT}} = -\sum_{(s_t, h_{t}^{'})\in \mathcal{D}_{\text{sft}}} \log \pi_\theta(h_t^{'} \mid s_t),
\end{equation}
where $\pi_\theta(\cdot)$ denotes the Planner model. 
This warm-up stage enables the Planner to imitate human planning behavior, learn common strategic patterns from high-quality dialogues, and acquire strong strategic priors.

\subsubsection{\textbf{Entropy-regularized RL Tuning}}
After warm-up fine-tuning with SFT, we perform an entropy-regularized RL tuning stage to enable more effective policy exploration. Specifically, we adopt a \textit{policy gradient update} method, updating the Planner at the end of each conversation using an entropy-regularized objective:
\begin{equation}\small 
\theta \leftarrow \theta - \alpha \, [\underbrace{\nabla_\theta \sum\limits_{t} \log \pi_\theta(h_t \mid s_t) \cdot R_t}_{\textit{(1)}} + \beta \underbrace{\nabla_\theta \, \frac{1}{T} \sum\limits_{t}  \mathrm{H}(\pi_\theta(\cdot \mid s_t))}_{\textit{(2)}}]. 
\label{eq:policy_update}
\end{equation}
Here, term (1) corresponds to the standard policy gradient, where $R_t$ denotes the cumulative reward. Term (2) represents an entropy regularization component, which maximizes the entropy of $\pi_\theta$ to prevent entropy collapse~\cite{cui2025entropy}, 
thereby mitigating the reduction of exploration diversity and preserving effective exploration during training. The $\alpha$ is the learning rate, and the $\beta$ is a coefficient to balance the regularization term. 

\vspace{+5pt}
\noindent\textbf{\textit{Reward Evaluation.}} The cumulative reward $R_t$ measures the total reward accumulating from turn $t$ to the final turn $T$ and is computed as follows:
\begin{equation}
   R_t = \sum_{i=0}^{T-t} \gamma^{i} r_{t+i}, 
\end{equation}
where $r_{t+i}$ denotes the reward at step $t+i$, and $\gamma$ is a discount factor that balances immediate and future rewards.

\textit{Turn-level Reward Computation.}
For each turn $t$, we adopt an LLM-as-a-judge approach, following prior work on reward modeling~\cite{deng2024plug}, to compute the turn-level reward $r_t$. Specifically, the Reward Expert is prompted with current conversational state $s_{t+1}$ (after taking action $a_t$) and a reward-specific evaluation instruction $\textit{prompt}_{\textit{rwd}}$, and produces a scalar score reflecting user satisfaction and task success. The process is formulated as:
\begin{equation}
r_{t} = \textit{Rewarder}(s_{t+1}, \textit{prompt}_{\textit{rwd}})
\end{equation}
This scalar score is normalized to a predefined range (e.g., $1$-$5$) to capture the quality of the actor's response. 
In addition to providing scalar rewards, the expert also determines whether the conversation should terminate. i.e., $d_t = \mathbb{I}[r_t > \tau]$, where $r_t$ exceeds a predefined threshold $\tau$ and $\mathbb{I}[\cdot]$ denotes the indicator function, generating a termination signal.

By maximizing these rewards during entropy-regularized RL tuning, the Planner learns to select macro-level strategies that drive both higher user satisfaction and improved recommendation success.

\section{Experimental Setup}
\label{sec:exp_setup}

\begin{table}[t]
\centering
\small
\caption{Statistics of our experimental datasets.}
\label{tab:dataset_stats}
\begin{tabular}{lccc}
\toprule
\textbf{Datasets} & \textbf{\#Dialogues} & \textbf{\#Utterances} & \textbf{\#Unique Items} \\
\midrule
\textbf{\textsc{Inspired}}~\cite{inspired} & 1{,}001  &  35{,}811  &  1{,}783 \\
\textbf{\textsc{ReDial}}~\cite{li2018towards}   & 10{,}006 & 182{,}150  & 51{,}699 \\
\bottomrule
\end{tabular}
\end{table}

\subsection{Datasets}
To evaluate our proposed method, we conduct experiments on two widely used multi-turn conversational recommendation datasets, summarized in Table~\ref{tab:dataset_stats}.

\textbf{\textsc{Inspired}}~\cite{inspired} is an English CRS dataset focused on movie recommendations. It is collected from human-human conversations in natural settings, where one participant plays the role of a recommender and the other acts as a user. Importantly, each dialogue turn is annotated with high-level strategies grounded in social science theories, ensuring that the annotated strategies reflect realistic conversational behaviors. This makes \textsc{Inspired} particularly suitable for evaluating strategy-aware CRS models.

\textbf{\textsc{ReDial}}~\cite{li2018towards} is another English CRS dataset for movie recommendations, constructed via crowd-sourcing on Amazon Mechanical Turk (AMT). It contains multi-turn dialogues between two assigned roles: a \emph{seeker}, who requests recommendations, and a \emph{recommender}, who provides them. Unlike \textsc{Inspired}, which emphasizes strategy annotation, \textsc{ReDial} focuses on content-level annotations, such as movie mentions and user preferences. The dialogues combine three complementary facets of conversational recommendation: goal-directed recommendation, open-domain chit-chat, and question-answering. This integration of multiple interaction types reflects realistic conversational dynamics in real-world recommendation scenarios.

\begin{table}[h]
\centering
\setlength{\tabcolsep}{4pt} 
\caption{Comparison of different baseline methods.}
\label{tab:baseline_comparison}
\begin{tabular}{lcccc}
\toprule
\textbf{Methods} & \textbf{Learnable} & \makecell{\textbf{Preference} \\ \textbf{Reasoning}} & \makecell{\textbf{Fact} \\ \textbf{Grounding}} & \makecell{\textbf{Strategy} \\ \textbf{Optimization}} \\
\midrule
\textbf{UniCRS}~\cite{Wang_2022unicrs}      & \cmark & \xmark & \cmark & \xmark \\
\textbf{BARCOR}~\cite{zhou2022barcor}      & \cmark & \xmark & \cmark & \xmark \\
\textbf{QwenCRS}~\cite{qwen2.5} & \xmark & \xmark & \xmark & \xmark \\
\textbf{LlamaCRS}~\cite{llama3} & \xmark & \xmark & \xmark & \xmark \\
\textbf{InterCRS}~\cite{wang2021interact}       & \xmark  & \xmark & \xmark & \xmark \\
\textbf{PCCRS}~\cite{pccrs}       & \xmark & \xmark & \cmark & \cmark \\
\textbf{RSO (Ours)}      & \cmark & \cmark & \cmark & \cmark \\
\bottomrule
\end{tabular}
\end{table}

\subsection{Baselines}
We compare our RSO method with a set of representative baselines, including both state-of-the-art PLM-based methods (UniCRS~\cite{Wang_2022unicrs} and BARCOR~\cite{wang2022barcor}) and recent LLM-based conversational recommenders (QwenCRS~\cite{qwen2.5}, LlamaCRS~\cite{llama3.1}, InterCRS~\cite{wang2023rethinking}, and PC-CRS~\cite{pccrs}). 

\begin{itemize}[leftmargin=*]
    \item \textbf{UniCRS}~\cite{Wang_2022unicrs}: A unified CRS model based on knowledge-enhanced prompt learning, which integrates recommendation and conversation subtasks within a knowledge-enhanced prompt learning framework. It incorporates fused knowledge, task-specific tokens, and dialogue context into prompt design to enable seamless subtask interaction and improve CRS performance.
    \item \textbf{BARCOR}~\cite{wang2022barcor}: A unified CRS framework that jointly tackles the recommendation and conversation tasks with a single model. This framework leverages BART as the backbone and incorporates a lightweight movie-domain knowledge graph to enhance the factual grounding.
    \item \textbf{QwenCRS}~\cite{qwen25}: A CRS method implemented with \texttt{Qwen2.5-7B-Instruct}, which is adapted to the task via in-context learning. It leverages the model’s strong in-context learning ability to handle conversational recommendations without additional fine-tuning.
    \item \textbf{LlamaCRS}~\cite{llama3}:  A CRS method built on top of \texttt{Llama3.1-8B-Instruct}, which is adapted to the task via prompting. It utilizes the model’s instruction-following capability to perform CRS.
    \item \textbf{InterCRS}~\cite{wang2023rethinking}: A session-level CRS that interleaves recommendation with natural conversation and models the interaction flow using a Transformer-based architecture.
    \item \textbf{PCCRS}~\cite{pccrs}: A recent LLM-based framework that explicitly balances persuasiveness and credibility in CRS by post-hoc self-reflection and strategy conditioning.
\end{itemize}
Table~\ref{tab:baseline_comparison} summarizes their capabilities along four key dimensions: \emph{learnability}, \emph{preference reasoning}, \emph{fact grounding}, and \emph{strategy optimization}. 
Most existing baselines are learnable but lack explicit modules for user preference reasoning or factual grounding, and they generally do not optimize macro-level strategies. 
In contrast, our RSO framework is the only approach that simultaneously supports learnability, dynamically infers user preferences, incorporates factual grounding, and optimizes strategy planning.

\begin{table*}[t]
\centering
\caption{Performance comparison of different CRS methods on the \textsc{INSPIRED} and \textsc{ReDial} datasets. We report results on Watching Intention (WI), Persuasiveness (PRS), Credibility (Cred), Conversation Success Rate (Conv-SR), and Distinct-2 (Dist-2). Best values are \textbf{bolded}; second best are \underline{underlined}.}
\label{tab:results_crs}
\resizebox{\textwidth}{!}{
\begin{tabular}{@{\hspace{1em}} l 
                *{5}{c} 
                @{\hspace{2em}}
                *{5}{c} 
                @{\hspace{1em}}}
\toprule
\textbf{Datasets ($\rightarrow$)} & \multicolumn{5}{c}{\textbf{\textsc{Inspired}}} & \multicolumn{5}{c}{\textbf{\textsc{ReDial}}} \\
\cmidrule(lr){2-6} \cmidrule(l){7-11}
\textbf{Methods ($\downarrow$)} & 
\textbf{WI} & 
\textbf{PRS} & 
\textbf{Cred} & 
\textbf{Conv-SR} & 
\textbf{Dist-2} &
\textbf{WI} & 
\textbf{PRS} & 
\textbf{Cred} & 
\textbf{Conv-SR} & 
\textbf{Dist-2} \\
\midrule
\textbf{UniCRS}          & 2.1010            & 0.8542            & 1.1187            & 0.2121            & 0.7567                & 2.2879            & 0.7059            & 0.8981            & 0.3838            & 0.7909 \\
\textbf{BARCOR}          & 2.5758            & 0.8654            & 1.5294            & 0.4848            & \underline{0.9861}                & 2.5000            & 0.8593            & 2.0187            & 0.4949            & \underline{0.9941} \\
\textbf{QwenCRS}    & 3.5556            & 0.9667            & 0.0000            & \underline{0.9596}            & 0.9160                & 3.3081            & 0.8452            & 0.2871            & \underline{0.9444}& 0.9170 \\
\textbf{LlamaCRS}   & \underline{3.6768}& \underline{0.9778}& 1.1313            & 0.7475            & 0.9427                & \underline{3.6869}            & \textbf{0.9821}            & 1.2190            & 0.8232            & 0.9352 \\
\textbf{InterCRS} & 3.5354            & \textbf{1.0000}   & 1.2429            & 0.8081            & 0.9564                & 3.6616            & 0.9750& 1.2797            & 0.8283            & 0.9163 \\
\textbf{PCCRS}           & 3.5152            & 0.9231            & \underline{3.2884}& 0.8889            & 0.8571                & 3.5152            & 0.9231            & \underline{3.2560}& 0.8889            & 0.8381 \\
\textbf{RSO (Ours)}            & \textbf{3.8788}   & \textbf{1.0000}   & \textbf{3.8333}   & \textbf{0.9798}   & \textbf{0.9979}       & \textbf{3.8393}   & \underline{0.9783}   & \textbf{3.6271}   & \textbf{0.9798}   & \textbf{0.9974} \\
\bottomrule
\end{tabular}
}
\end{table*}

\subsection{Evaluation Metrics}
Following~\cite{pccrs,Wang_2022unicrs}, we evaluate multiple aspects of CRS performance using the following metrics, including:
\begin{itemize}[leftmargin=*]
\item \textit{Conversation Success Rate (\textbf{Conv-SR})} quantifies the proportion of dialogues in which the user ultimately accepts the recommended item. 
The Conv-SR is based on the reward score assigned by the LLM-based evaluator.
\item \textit{Watching Intention (\textbf{WI})} quantifies the user’s willingness to accept the recommendation after completing the full interaction with the system, serving as an indicator of the ultimate effectiveness of the recommendation in shaping user intent.
\item \textit{Persuasiveness (\textbf{PRS})} evaluates the extent to which the system’s explanations influence the user’s intention to watch the recommended item. Inspired by human studies of persuasion~\cite{persuasion}, \textit{PRS} is defined as:
\[
P = 1 - \frac{i_{\text{true}} - i_{\text{post}}}{i_{\text{true}} - i_{\text{pre}}},
\]
where $i_{\text{pre}}$ is the user’s initial intention, $i_{\text{post}}$ is the final intention after completing the conversation, and $i_{\text{true}}$ is the user's real intention regarding the recommended item, which is calculated by providing all the information of the recommended item to the user. To ensure rationality, we impose the constraint $i_{\text{true}} \geq i_{\text{post}}$. The score $P \in [0,1]$, with higher values indicating stronger persuasion capability.  
\item \textit{Credibility (\textbf{Cred})} measures the degree to which the recommender’s utterances are consistent with factual information.  
\item \textit{Distinct-2 (\textbf{Dist-2})} measures the lexical diversity of generated responses, calculated as the ratio of unique bigrams to the total number of bigrams in the output. 
Higher Dist-2 scores indicate more varied utterances, whereas lower scores suggest redundancy or template-like responses.
\end{itemize}

Conversational quality metrics are automatically evaluated under the \textit{LLM-as-a-Judge} paradigm, using \texttt{Qwen2.5-7B-Instruct} with fine-grained scoring rubrics, while Dist-2 is directly computed from the generated outputs. 
To ensure fairness, all models are assessed under identical prompting and generation settings.
In addition to evaluating the effectiveness of CRS strategy optimization, we also measure the accuracy of recommendation outcomes. Following~\cite{pccrs}, we adopt the following metrics, including:
\begin{itemize}[leftmargin=*]
\item \textit{Recall@$k$ (\textbf{R@$k$})} evaluates whether the ground-truth item appears among the top-$k$ recommended candidates. We report Recall@$1$ and Recall@$5$.  
\item \textit{Recommendation Success Rate (\textbf{Rec-SR})} measures the proportion of dialogues in which the recommender successfully recommends the target item.
\end{itemize}

\begin{table*}[t]
    \centering
    \caption{Comparative performance on recommendation accuracy across \textsc{INSPIRED} and \textsc{ReDial}. We report Recommendation Success Rate (Rec-SR) and Recall@$k$ ($k=1,5$). \textit{Improve.} denotes relative improvement: percentages (\%) for gains below 100\%, and multiplicative factors ($\times$) for gains above 100\%, compared with the UniCRS baseline.}
    \label{tab:rec_success}
    \begin{tabular}{
>{\raggedright\arraybackslash}p{2cm}
>{\centering\arraybackslash}p{1.5cm}
>{\centering\arraybackslash}p{1.5cm}
>{\centering\arraybackslash}p{1.5cm}
>{\centering\arraybackslash}p{1.5cm}
>{\centering\arraybackslash}p{1.5cm}
>{\centering\arraybackslash}p{1.5cm}
>{\centering\arraybackslash}p{1.5cm}
>{\centering\arraybackslash}p{1.5cm}}
    \toprule
     \textbf{Datasets ($\rightarrow$)} & \multicolumn{4}{c}{\textbf{\textsc{Inspired}}}    & \multicolumn{4}{c}{\textbf{\textsc{ReDial}}} \\ \cmidrule(lr){2-5} \cmidrule(l){6-9} 
    \textbf{Methods ($\downarrow$)}   & \textbf{Rec-SR} & \textbf{Recall@1} & \textbf{Recall@5} & \textit{Improve.} $\uparrow$ & \textbf{Rec-SR} & \textbf{Recall@1} & \textbf{Recall@5} & \textit{Improve.} $\uparrow$\\ 
    \midrule
    \textbf{UniCRS}      & 0.172   & 0.556    & \underline{0.838} & -     & 0.015 & \underline{0.652} & 0.788        & -      \\
    \textbf{BARCOR}      & 0.424   & 0.121    & 0.646    & 15.1\% & 0.354 & 0.495 & 0.707  & 7.4$\times$     \\
    \textbf{QwenCRS}     & 0.384   & 0.354    & 0.687    & 23.0\% & 0.308 & 0.308 & 0.652  & 6.3$\times$     \\
    \textbf{LlamaCRS}    & \textbf{0.576}   & 0.475    & 0.778    & \underline{71.1\%} & \underline{0.450} & 0.384 & 0.647        & 9.5$\times$     \\
    \textbf{InterCRS}    & 0.323   & 0.485    & 0.758    & 21.8\% & 0.222 & 0.384 & 0.677  & 4.4$\times$     \\
    \textbf{PCCRS}       & 0.525   & \underline{0.586} & 0.768    & 67.4\% & \textbf{0.566} & 0.596 & \underline{0.833}        & \underline{12.2$\times$}     \\

    \textbf{RSO (Ours)}& \underline{0.535} & \textbf{0.758}    & \textbf{0.889}    & \textbf{84.5\%} & \textbf{0.566} & \textbf{0.742} & \textbf{0.884}  & \textbf{12.3$\times$}     \\
    \bottomrule
    \end{tabular}
\end{table*}

\subsection{Implementation Details}
Our framework is implemented on top of \texttt{Qwen2.5-7B-Instruct}, which serves as the backbone LLM for all experts, except for the \textit{Planner}, where we employ \texttt{RoBERTa}~\cite{liu2019roberta} as the backbone model. Each expert (\textit{Planner}, \textit{Preference Reasoner}, \textit{Factual Retriever}, and \textit{Actor}) is instantiated via prompting. The Factual Retriever retrieves knowledge from the knowledge graph provided in the datasets, which is a curated subset of Wikipedia, ensuring factual grounding for recommendation. We set the maximum dialogue length to 10 turns; if a user does not accept a recommendation earlier, the conversation is automatically terminated. 
The reward model in our RSO framework outputs scores on a scale from 1 to 5, designed for the quality of the conversation. During training, we first perform SFT on \textsc{Inspired} to initialize the Planner, and then apply entropy-regularized RL tuning with the reward model as feedback. This two-stage setup stabilizes optimization and prevents direct RL from causing divergence. For all experiments, we train models on both \textsc{Inspired} and \textsc{ReDial}, which are in the movie recommendation domain and share similar content, ensuring consistency and transferability.  

We adopt the AdamW optimizer with a batch size of 16 and linear learning rate decay. The learning rates are set to $6 \times 10^{-6}$ for SFT and $1 \times 10^{-4}$ for reinforcement learning. The SFT stage is trained for 10 epochs, while the entropy-regularized RL tuning stage is optimized for 10 epochs. Each epoch consists of 100 dialogue generations with a maximum of 10 turns. The discount factor is set to $\gamma=0.99$, and rewards are scaled to the range $[0,1]$. To reduce variance in reward estimation, we follow \cite{wang2023reward_average} by sampling each reward computation 10 times and averaging the results to obtain the final reward value for each response. All experiments are conducted on 1 $\times$ A100 GPUs (80GB).

\section{Experimental Results}
\label{sec:exp_results}
\subsection{Main Results}

Table~\ref{tab:results_crs} presents the main comparison across two benchmark datasets, using five CRS evaluation metrics. To investigate the impact of strategy optimization in our RSO framework, we make the following key observations.

First, LLM-based models (QwenCRS, LlamaCRS, InterCRS, PCCRS, and our RSO) consistently outperform PLM-based methods (UniCRS and BARCOR) across nearly all metrics and datasets. This performance gap highlights the advantage of leveraging LLMs’ stronger reasoning and generation capabilities for conversational recommendation. For example, UniCRS and BARCOR yield notably weaker results in WI and Conv-SR, indicating their limited ability to sustain user engagement and deliver effective recommendations compared to LLM-based approaches.

Second, our proposed RSO framework achieves the best or highly competitive results across all metrics on both datasets. In particular, RSO attains the highest WI, PRS, Credibility, and Conv-SR simultaneously, while maintaining strong diversity (Dist-2). On the \textsc{Inspired} dataset, RSO substantially surpasses PCCRS in both Credibility and WI on \textsc{ReDial}. RSO also achieves better performance in PRS and Conv-SR, while delivering dramatically higher Credibility compared to QwenCRS. These findings demonstrate that our reinforced strategy optimization framework not only enhances persuasiveness, but also ensures credibility.

Finally, while PCCRS is a strong LLM-based baseline that incorporates strategies through the static prompts, its improvements remain limited due to the absence of explicit optimization over the strategy space. In contrast, our RSO framework explicitly optimizes strategy planning via the Planner expert, supported by Preference Reasoner and Factual Retriever experts. This disentangled design leads to consistent improvements across all dimensions. The improvements over PCCRS confirm the effectiveness of performing reinforced strategy optimization.

\begin{figure*}[htbp]
    \centering
    \begin{subfigure}{0.48\textwidth}
        \centering
        \includegraphics[width=\linewidth]{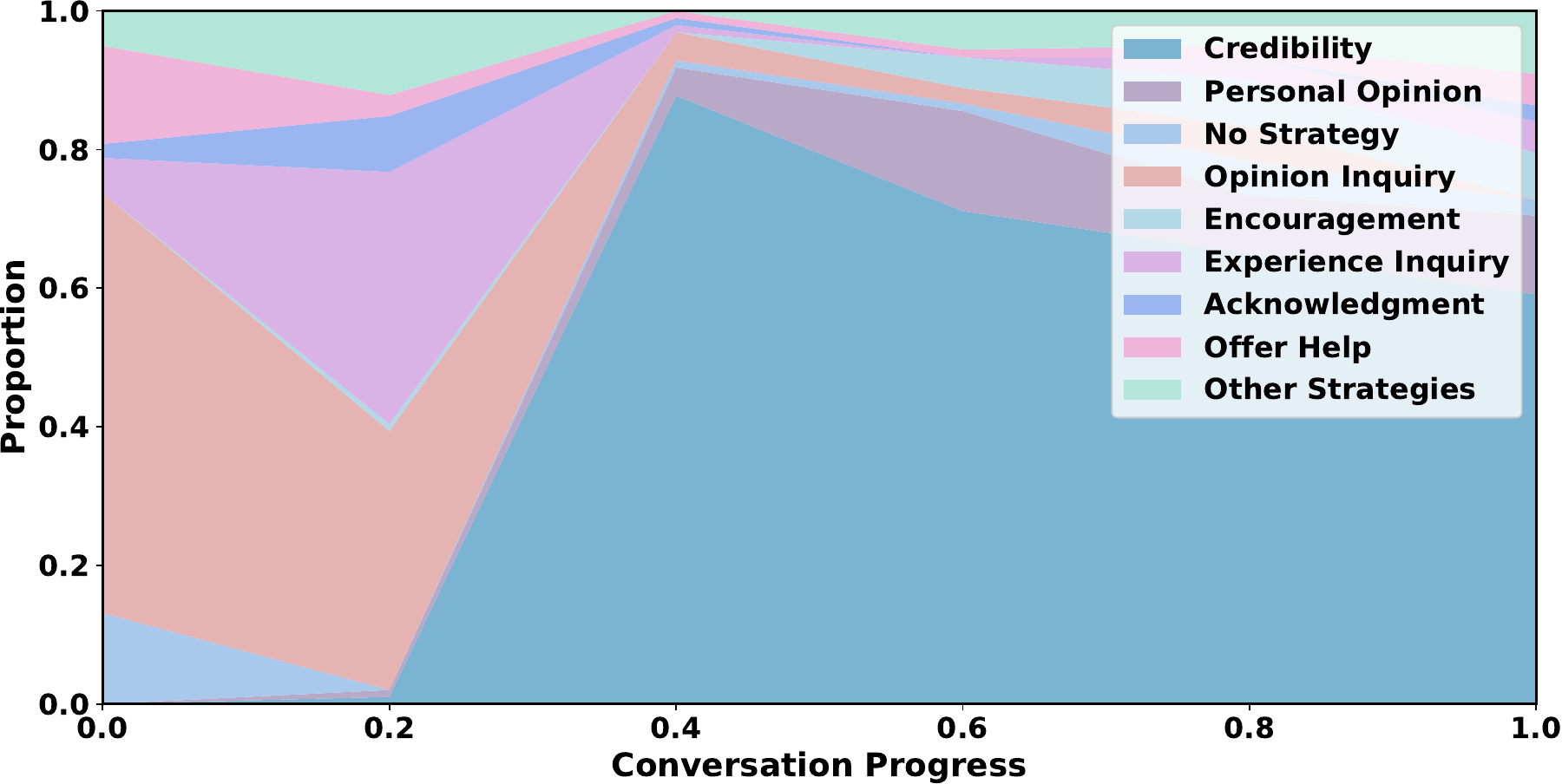}
        \caption{Strategy distribution across conversation progress before entropy-regularized RL tuning on the \textsc{Inspired} dataset. }
        \label{}
    \end{subfigure}
    \hfill
    \begin{subfigure}{0.48\textwidth}
        \centering
        \includegraphics[width=\linewidth]{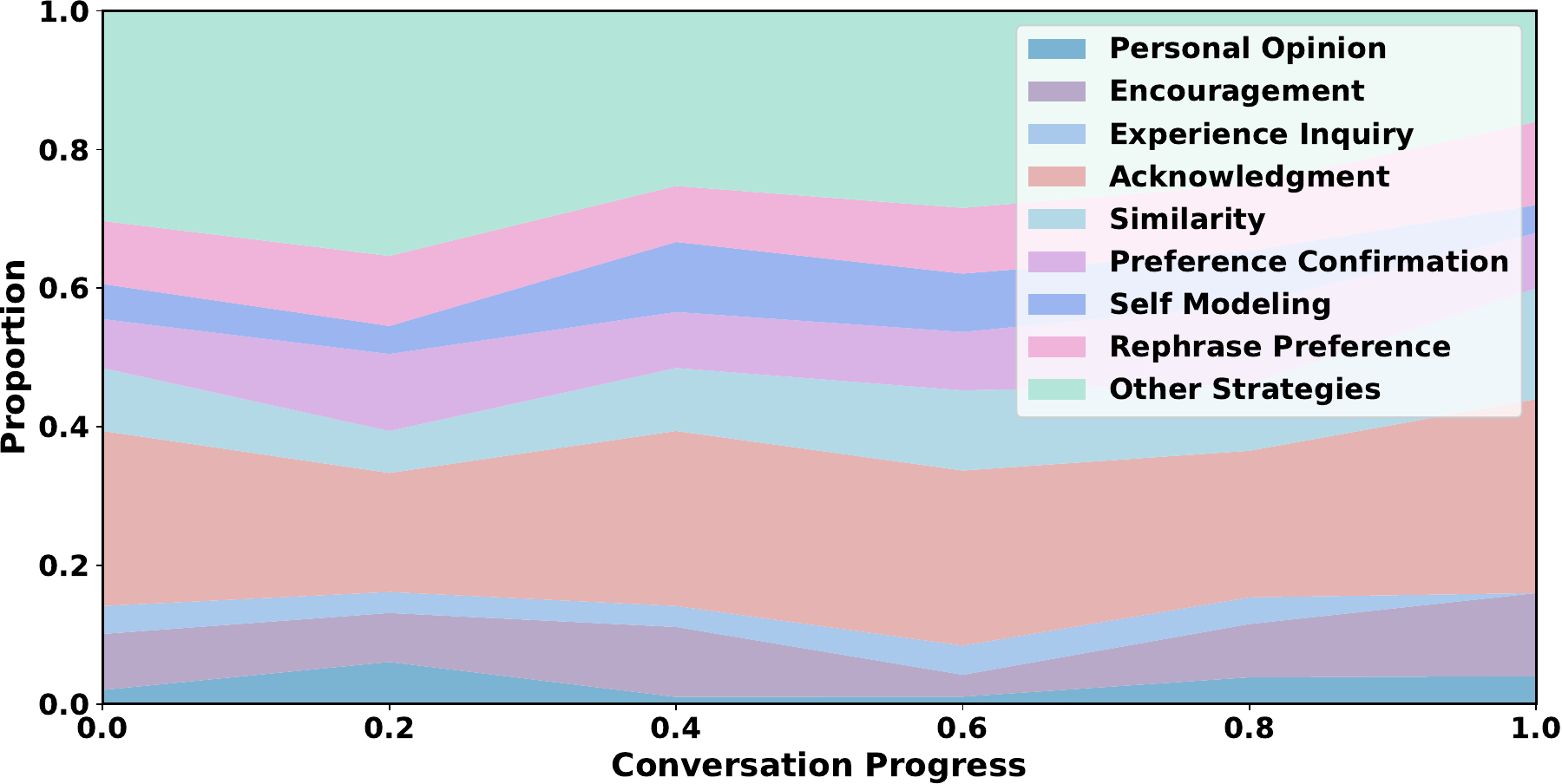}
        \caption{Strategy distribution across conversation progress after RL optimization with entropy-regularized policy updates.}
        \label{}
    \end{subfigure}

    \caption{Comparison of macro-level strategy distributions  across conversation progress before (a) and after (b) Entropy-regularized RL tuning on the \textsc{Inspired} dataset. Entropy-regularized RL reduces over-reliance on a few dominant strategies and encourages broader, context-aware strategy exploration, resulting in more adaptive and balanced conversational behaviors.}
    \label{fig:strategy-distribution}
\end{figure*}


    
    


\begin{figure}
    \centering
    \includegraphics[width=\linewidth]{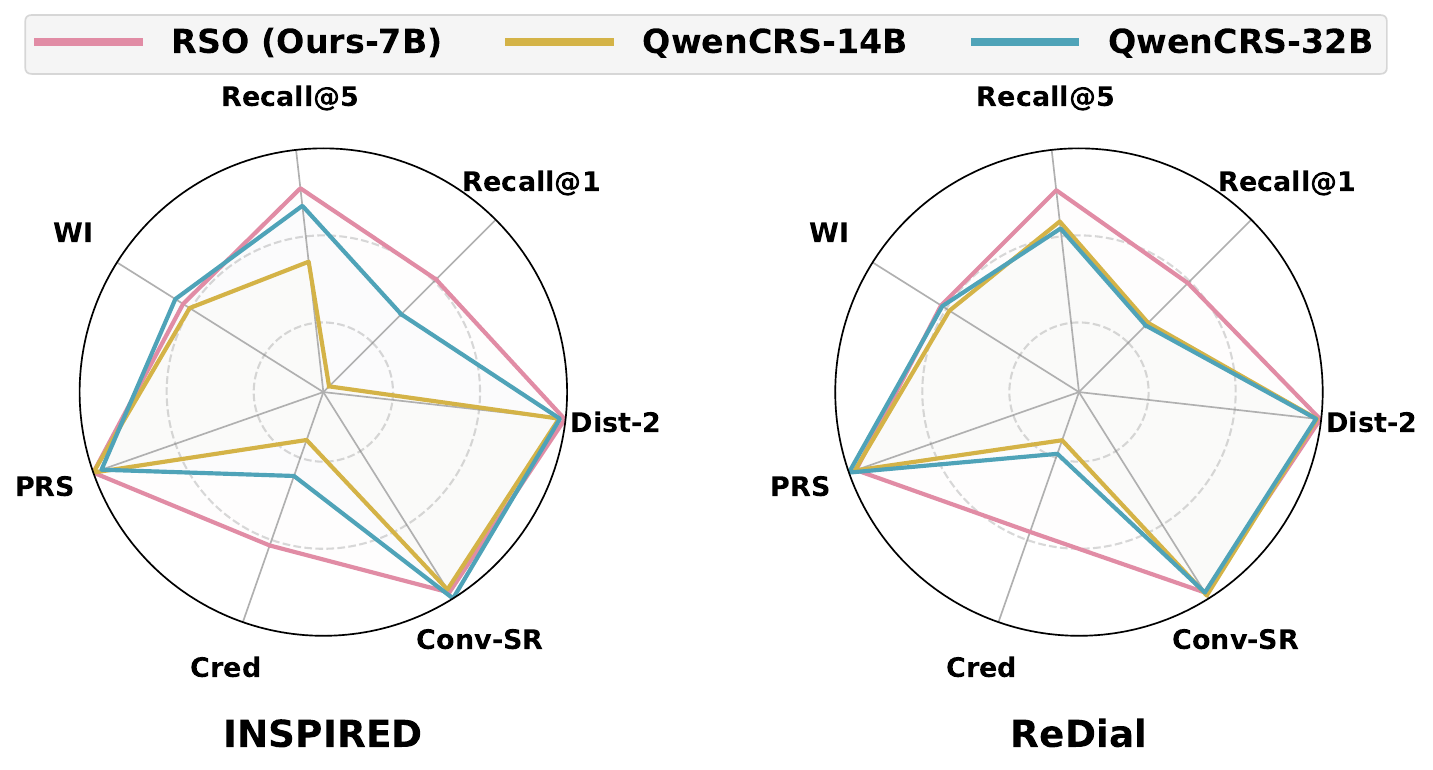}
    \caption{Comparison of our RSO (7B) with QwenCRS-14B and QwenCRS-32B on \textsc{Inspired} and \textsc{ReDial} datasets. }
    \label{fig:14b_32b_results}
\end{figure}
\subsection{In-depth Analysis}


\subsubsection{\textbf{Impact of Reinforced Strategy Optimization on CRS performance}}

\textit{How do macro-level strategy planning and micro-level strategy adaptation jointly affect CRS performance?} Reinforced macro-level strategy planning learns to select context-appropriate strategies, while micro-level strategy adaptation personalizes and grounds each strategy execution in factual knowledge, jointly improving overall interaction performance.

At the macro level, the Planner determines which type of conversational action to take (e.g., recommending, clarifying, or explaining). Without such planning, the system may issue recommendations prematurely or fail to probe user preferences adequately, resulting in low accuracy. By explicitly optimizing the Planner with reinforcement signals, our framework ensures that recommendations are made at the right time and under the right conversational context, thereby improving Conv-SR.

At the micro level, expert modules coordinate to adapt the selected strategy to user-specific contexts: the Preference Reasoner dynamically infers evolving user preferences, while the Fact Retriever provides factual grounding to keep responses relevant and accurate. Together, they ensure that the recommended candidates are both relevant to the user’s evolving preferences and factually trustworthy.

The hierarchical decomposition disentangles macro-level strategy planning from micro-level execution. Macro-level planning focuses on strategy planning, while micro-level adaptation guarantees that each chosen strategy is instantiated effectively for the specific user. As shown in Table~\ref{tab:rec_success}, this complementary design enhances recommendation performance.

\subsubsection{\textbf{Impact of Entropy-regularized RL Tuning on Strategy Distribution}}
Fig.~\ref{fig:strategy-distribution} illustrates how our entropy-regularized RL tuning reshapes the distribution of macro-level strategies throughout a conversation on the \textsc{Inspired} dataset.
With only SFT (Fig.~\ref{fig:strategy-distribution}a), the Planner exhibits a strong bias toward a few dominant strategies, such as \textit{Credibility} and \textit{Opinion Inquiry}, resulting in repetitive and less adaptive conversations. Many valuable strategies—like \textit{Preference Confirmation}, \textit{Rephrase Preference}, and \textit{Self Modeling}—are rarely activated, which limits the system’s ability to dynamically adapt to evolving user needs.

To address this, we introduce entropy-regularized RL tuning (Fig.~\ref{fig:strategy-distribution}b), as defined in Eq.~\ref{eq:policy_update}, which explicitly maintains exploration diversity and prevents premature convergence to a small set of strategies. As a result, previously under-utilized strategies (e.g., \textit{Rephrase Preference}, \textit{Preference Confirmation}, \textit{Similarity}) are invoked more frequently across dialogue turns. This shift shows that entropy-regularized RL tuning encourages broader exploration of the strategy space, enabling the Planner to discover more contextually appropriate strategies rather than over-relying on “safe” defaults.

Overall, entropy-regularized RL tuning mitigates strategy collapse and promotes a more adaptive mixture of behaviors, leading to better-timed, more persuasive, and user-aligned dialogues—ultimately improving both conversational quality and recommendation success.

\subsubsection{\textbf{Reinforced Strategy Optimization (7B) vs. Large-Scale Baselines (14B \& 32B)}}
As shown in Fig.~\ref{fig:14b_32b_results}, our RSO, built on a 7B \textit{network-of-experts} architecture, achieves performance that is comparable to the much larger QwenCRS-14B and QwenCRS-32B baselines across both \textsc{Inspired} and \textsc{ReDial}, and even surpasses them on several key metrics. 
The network-of-experts comprises a \textit{Planner} for macro-level strategy selection, an \textit{Actor} for response generation, and two auxiliary experts, namely a \textit{Preference Reasoner} and a \textit{Fact Retriever}, which provide dynamic user preference modeling and factual grounding, respectively. 
This design enables RSO to jointly optimize strategic planning and context adaptation, allowing it to match or exceed the performance of larger models with far fewer parameters. 
Notably, RSO achieves higher Recall@1 and Recall@5, indicating more accurate recommendation targeting, and substantially improves credibility (Cred), confirming its strength in generating factually grounded and trustworthy responses. 
These gains are accompanied by competitive performance on conversational success rate (Conv-SR), diversity (Dist-2), and engagement quality (WI), which shows that reinforced strategy optimization can enhance CRS performance efficiently without simply relying on model scale.

\begin{figure*}[htbp]
    \centering
    \begin{subfigure}{0.32\textwidth}
        \centering
        \includegraphics[width=\linewidth]{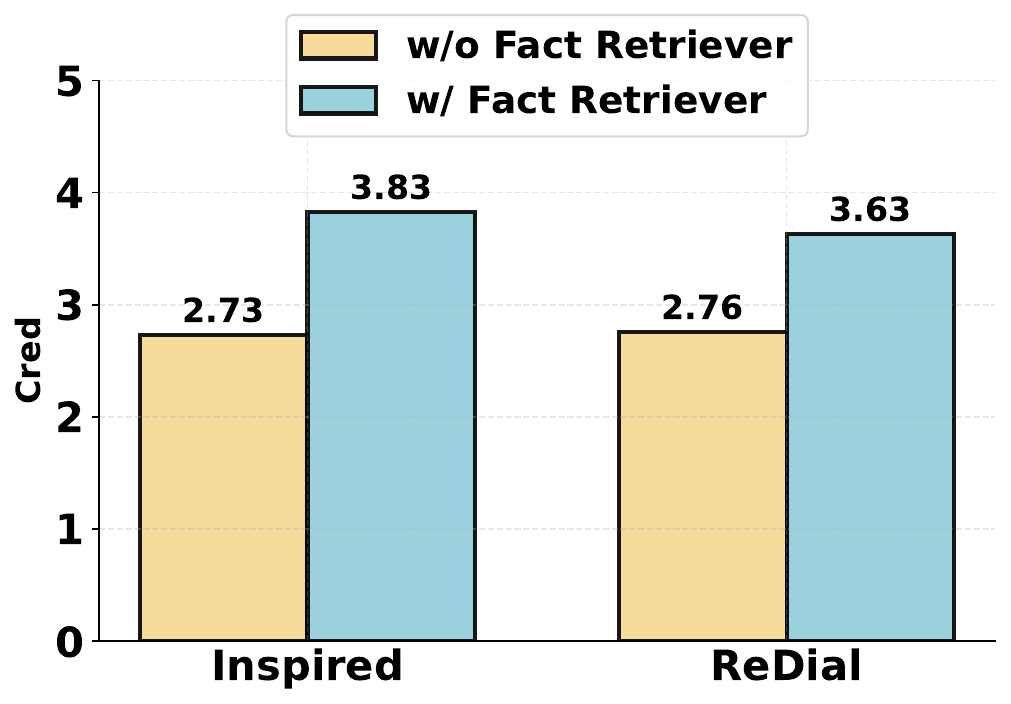}
        \caption{Effect of Fact Retriever}
        \label{fact_retriever_comparison}
    \end{subfigure}
    \hfill
    \begin{subfigure}{0.32\textwidth}
        \centering
        \includegraphics[width=\linewidth]{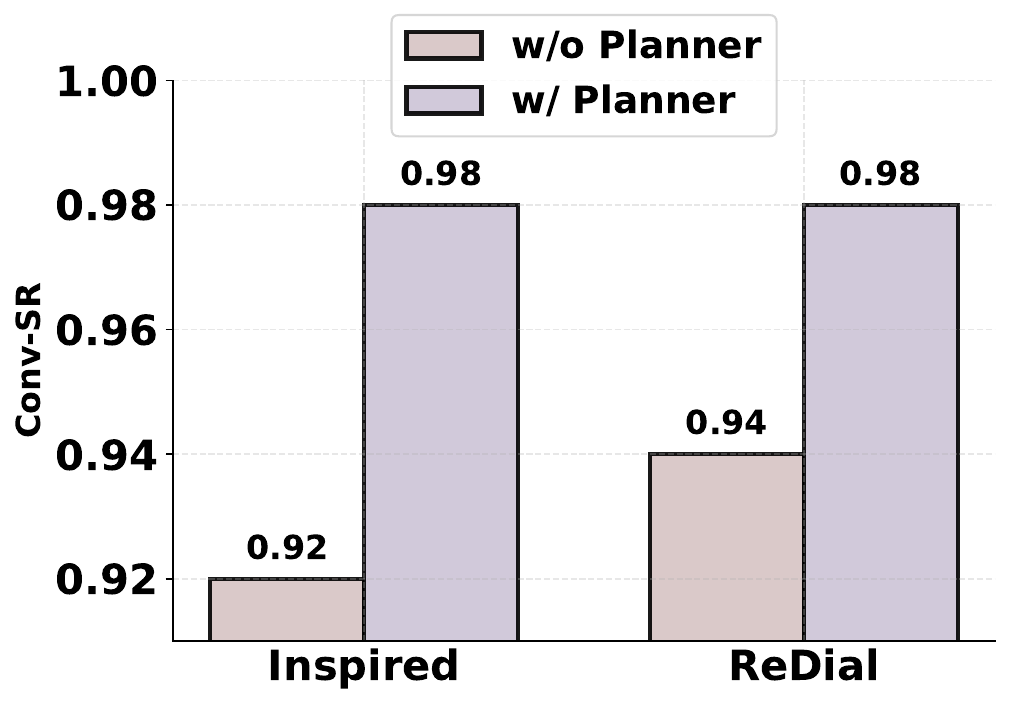}
        \caption{Effect of Planner}
        \label{planner_comparison}
    \end{subfigure}
    \hfill
    \begin{subfigure}{0.32\textwidth}
        \centering
        \includegraphics[width=\linewidth]{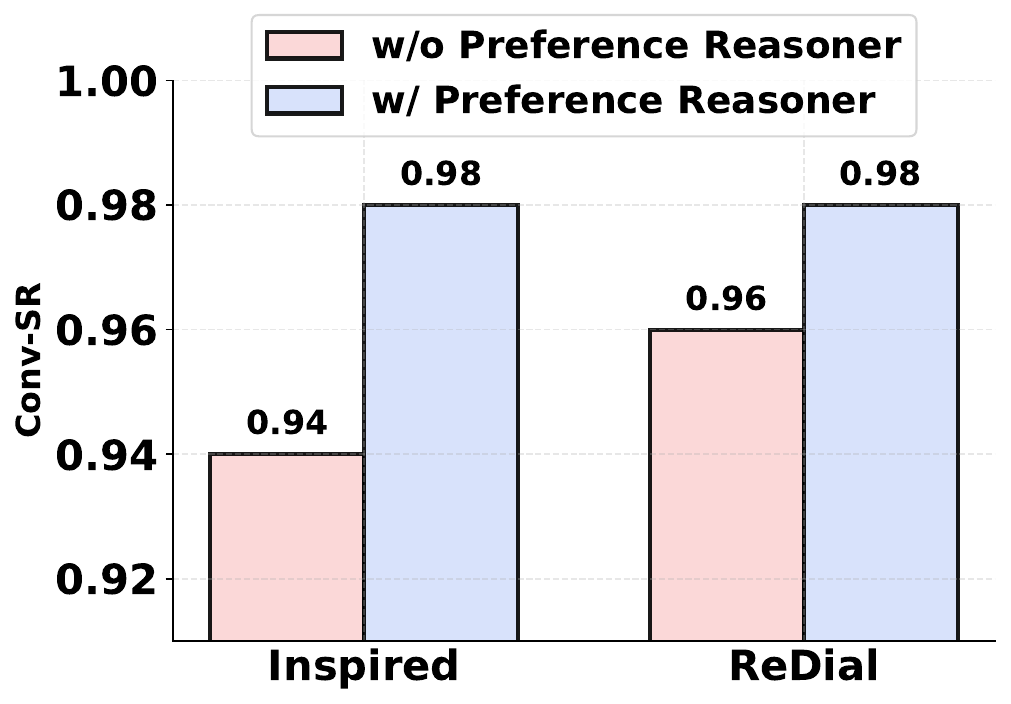}
        \caption{Effect of Preference Reasoner}
        \label{preference_reasoner_comp}
    \end{subfigure}
    \caption{Ablation studies on the impact of the Fact Retriever, Planner, and Preference Reasoner on \textsc{INSPIRED} and \textsc{ReDial} datasets. The Fact Retriever significantly improves \textit{Credibility}, while both the Planner and Preference Reasoner are essential for achieving high \textit{Conversation Success Rate}.}
    \label{ablation}
\end{figure*}

\subsection{Ablation Study}

\subsubsection{\textbf{Impact of Fact Retriever on RSO}} 
As shown in Fig.~\ref{fact_retriever_comparison}, incorporating the Fact Retriever substantially improves the credibility of system responses on both \textsc{Inspired} and \textsc{ReDial}. With fact retrieval, the credibility score increases from 2.73 to 3.83 on \textsc{Inspired}, and from 2.76 to 3.63 on \textsc{ReDial}, demonstrating consistent gains across datasets. 

The reason lies in how the Fact Retriever integrates factual grounding with user-specific reasoning. Based on dialogue history, the Preference Reasoner infers the user’s current intent and candidate items, which then guide retrieval from the knowledge graph (a curated subset of Wikipedia). By aligning retrieved evidence with inferred preferences, the system ensures factual grounding. Without fact retrieval, the system is more prone to hallucinations, leading to reduced user trust. With fact retrieval, the generated responses are both informative and trustworthy, thereby improving overall conversational effectiveness.

\subsubsection{\textbf{Impact of Planner on RSO}}
As shown in Fig.~\ref{planner_comparison}(b), the inclusion of the Planner leads to a substantial improvement in \emph{conversational success rate} (Conv-SR), from 0.92 to 0.98 on \textsc{Inspired} and from 0.94 to 0.98 on \textsc{ReDial}. This demonstrates that macro-level strategy planning plays a decisive role in achieving successful recommendations.

The Planner equips the CRS with explicit strategies — such as leveraging \textit{Similarity} to resonate with user preferences, sharing \textit{Personal Opinions} to enhance persuasiveness, and providing \textit{Factual Information} to build trust — enabling the system to actively guide conversations toward uncovering user intent and securing positive recommendation outcomes.

These results confirm that macro-level planning is essential for driving higher success rates, transforming CRSs into proactive, goal-directed systems that achieve CRS success.

\subsubsection{\textbf{Impact of Preference Reasoner on RSO}}
As shown in Fig.~\ref{preference_reasoner_comp}, removing the Preference Reasoner leads to a noticeable drop in recommendation accuracy. Without explicit preference reasoning, the Conv-SR falls from 0.98 to 0.94 on \textsc{Inspired}, and from 0.98 to 0.96 on \textsc{ReDial}. This indicates that the system struggles to converge on the correct recommendation when user preferences are not explicitly inferred and represented.

The underlying reason is that user preferences are often implicit and distributed across multiple conversational turns. If the model relies only on the current utterance, important contextual cues may be overlooked, leading to suboptimal strategy execution. By contrast, the Preference Reasoner explicitly synthesizes historical dialogue context into structured preference descriptions, making hidden signals more salient and accessible to downstream modules. 
These findings confirm the value of introducing explicit reasoning modules in CRS. Rather than treating user preferences as transient signals, our approach elevates them into structured representations that guide both strategy planning and response generation. The improvement in both efficiency and accuracy validates Preference Reasoner as a critical expert in our hierarchical NoE framework.

\subsection{Case Study}
\begin{figure}[t] 
    \centering
    \includegraphics[width=\columnwidth]{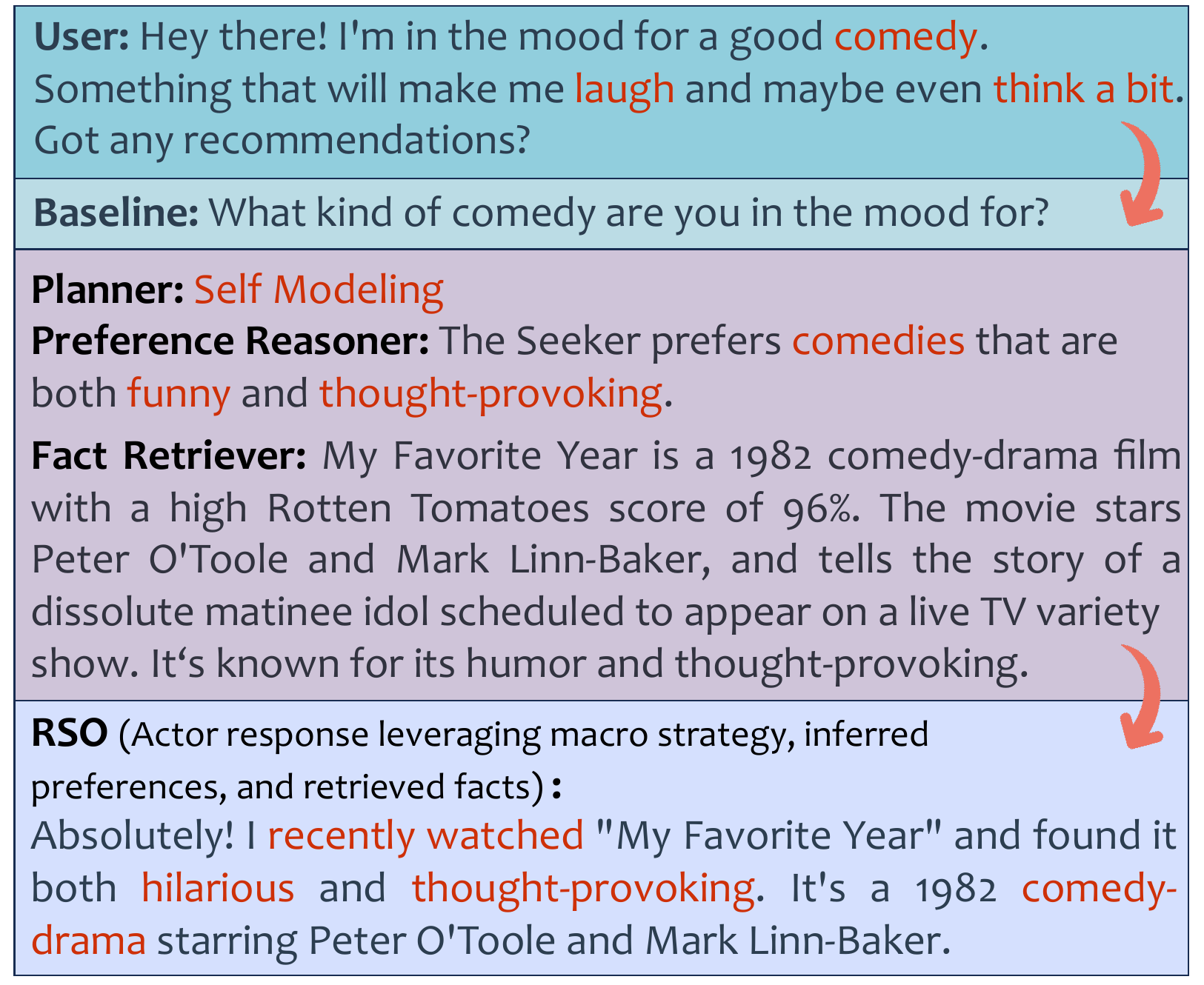}
    \caption{Qualitative case study on the \textsc{Inspired} dataset. Our RSO framework first selects a macro-level strategy, then grounds its response with inferred user preference and retrieved factual information about the recommended item.}
    \label{fig:case_study}
\end{figure}

To illustrate the effectiveness of our framework, Fig.~\ref{fig:case_study} compares our RSO framework (bottom) with a baseline LLM-based CRS without explicit strategy optimization (top).
The baseline system simply echoes the user query and asks a follow-up question (“What kind of comedy are you in the mood for?”), which increases the number of turns and leaves the burden on the user to further clarify their intent.

In contrast, our RSO framework first invokes the Planner to select a macro-level strategy (Self Modeling), then uses the Preference Reasoner to infer that the user prefers comedies that are both funny and thought-provoking. This inferred preference is passed to the Fact Retriever, which retrieves a relevant candidate (My Favorite Year, a 1982 comedy-drama with a 96\% Rotten Tomatoes score). Finally, the Actor integrates these components to produce a natural, engaging response that not only recommends the item but also justifies it with factual support (``I recently watched $\cdots$ hilarious and thought-provoking'').

By explicitly reasoning over user intent, grounding the response with supporting facts, and delivering a complete recommendation in a single turn, RSO achieves more personalized, credible, and persuasive recommendations while also being more efficient by avoiding unnecessary clarification turns. This case study highlights the key advantage of reinforced strategy optimization—RSO does not merely react to surface-level queries but actively interprets and dynamically satisfies implicit preferences.

\section{Related Work}

\subsection{PLM-based CRSs.}  
Conversational recommender systems (CRSs) allow users to interact with systems through free-form natural language to achieve recommendation goals \cite{li2018towards}. Early work primarily relied on pre-trained language models (PLMs) as backbones to generate human-like responses \cite{wang2022recindial}. These models supported user engagement by enabling interaction across three key stages: (1) preference elicitation through user input \cite{zhang2025reinforced,liu2025inference,lin2025igd}, (2) recommendation presentation \cite{sun2022enhancing,liu2025discrec}, and (3) user feedback \cite{wang2022barcor} that empowers users to influence the recommendation process directly. While effective in facilitating natural interaction, PLM-based approaches often faced limitations in capturing deeper user intent and providing adaptive strategies.

\subsection{LLM-based CRSs.}  
Large language models (LLMs) have demonstrated remarkable abilities~\cite{zhao2024comprehensive,zhang2025explicit,zhang2025learn,liu2025continual,zhou2025dreamdpo} in understanding user intents and generating context-aware responses.
With the advent of LLMs, CRS research has shifted from delivering factual information to actively persuading users, thereby increasing recommendation acceptance rates \cite{tran2024less,zhang2024text,collm}. LLMs provide greater flexibility to generate personalized content \cite{zhao2025nextquill,dpl,dep} and persuasive utterances through strategic reasoning \cite{zhao2025exploring}. However, prior studies also reveal that LLMs may inject misinformation in pursuit of persuasiveness, undermining user trust in the long run. These concerns resonate with social science and HCI findings, which emphasize the critical roles of persuasion and credibility in shaping user acceptance \cite{cialdini2004science,gkika2014persuasive}. Motivated by these insights, several works focus on enhancing persuasiveness \cite{slattery2020persuasion} and credibility \cite{yoo2010creating} in recommender systems.  

Despite these advances, LLM-based CRSs still face two key challenges. First, their performance heavily depends on internal parametric knowledge, leading to degradation in knowledge-scarce domains \cite{wang2023rethinking}. To alleviate this, recent studies~\cite{zhao2024pacar,huang2025recommender,xi2024memocrs} integrate LLMs with external agents or retrieval mechanisms for factual grounding. Second, while LLMs are capable of producing strategy-driven utterances~\cite{zhao2025exploring}, there is limited work on explicitly optimizing these strategies in conjunction with factual retrieval and user preference reasoning. This gap motivates our exploration of hierarchical strategy optimization that disentangles and integrates these components for more adaptive CRS.


\section{Conclusion}
In this work, we introduce the RSO framework, a reinforced strategy optimization framework for LLM-based CRSs.
By decomposing interaction decisions into macro-level strategy planning (via a Planner expert) and micro-level strategy adaptation (via an Actor and auxiliary experts for preference reasoning and factual grounding), RSO disentangles complex conversational subtasks into tractable components.
We further formulated macro-level strategy learning as a reinforcement learning problem guided by an LLM-based reward model, enabling automatic exploration of the strategy space.
Extensive experiments on two benchmark CRS datasets demonstrate that RSO achieves state-of-the-art results across both recommendation accuracy and conversational quality metrics, significantly improving overall user satisfaction.

\bibliographystyle{IEEEtran}
\bibliography{references}

\begin{IEEEbiographynophoto}{Xiaoyan Zhao} is currently a Ph.D. student in the Department of Systems Engineering and Engineering Management at The Chinese University of Hong Kong. Her research interests include LLM personalization and Agentic AI.

\par
\noindent\textbf{Ming Yan} obtained his bachelor's degree from the University of Science and Technology of China (USTC). His research interests include LLM personalization and LLM-based recommendation.

\par
\noindent\textbf{Yang Zhang} is a Research Fellow at the National University of Singapore. He obtained his Ph.D. from the University of Science and Technology of China (USTC). His research interests lie in recommender systems. He has authored over ten publications in top conferences and journals such as SIGIR and TKDE, including one that received the Best Paper Honorable Mention at SIGIR 2021. He has served as a program committee member and reviewer for top conferences and journals, including KDD, SIGIR, TKDE, and TOIS.

\par
\noindent\textbf{Yang Deng} is an Assistant Professor at Singapore Management University. He received the B.S. degree from the Beijing University of Posts and Telecommunications, the M.S. degree from Peking University, and the Ph.D. degree from The Chinese University of Hong Kong. He was previously a Postdoctoral Research Fellow with NExT++, School of Computing, National University of Singapore. His research interests include natural language processing, information retrieval, and deep learning.

\par
\noindent\textbf{Jian Wang} is a Postdoctoral Fellow at The Hong Kong Polytechnic University. He obtained his M.Sc. and B.Eng. degrees from the School of Computer Science and Engineering, South China University of Technology. His research interests include LLM-powered agents and conversational AI.

\par
\noindent\textbf{Fengbin Zhu} is a Postdoctoral Research Fellow at the National University of Singapore (NUS). He obtained his Ph.D. from NUS in June 2024. His research interests include multimodal and heterogeneous retrieval as well as AI for finance.

\par
\noindent\textbf{Yilun Qiu} is a master's student at the National University of Singapore (NUS). He obtained his bachelor's degree from the Southern University of Science and Technology. His research interest is LLM personalization.

\par
\noindent\textbf{Hong Cheng} is a Professor in the Department of Systems Engineering and Engineering Management at The Chinese University of Hong Kong. She received her Ph.D. from the University of Illinois at Urbana-Champaign in 2008. Her research interests include data mining, database systems, and machine learning. She received research paper awards at ICDE 2007, SIGKDD 2006, and SIGKDD 2005, as well as a certificate of recognition for the 2009 SIGKDD Doctoral Dissertation Award. She is also a recipient of the 2010 Vice-Chancellor’s Exemplary Teaching Award at CUHK.

\par
\noindent\textbf{Tat-Seng Chua} is the KITHCT Chair Professor at the School of Computing, National University of Singapore. He was the Acting and Founding Dean of the School from 1998 to 2000. His main research interests are in multimedia information retrieval and social media analytics, focusing on the extraction, retrieval, and question answering of text and rich media from the Web and social networks. He is the co-Director of NExT, a joint center between NUS and Tsinghua University that develops technologies for live social media search. Dr. Chua was the recipient of the 2015 ACM SIGMM Award for Outstanding Technical Contributions to Multimedia Computing, Communications, and Applications. He is the Chair of the steering committees of the ACM International Conference on Multimedia Retrieval (ICMR) and the Multimedia Modeling (MMM) conference series. He has served as General Co-Chair for ACM Multimedia 2005, ACM CIVR (now ACM ICMR) 2005, ACM SIGIR 2008, and ACM Web Science 2015. He serves on the editorial boards of four international journals and is the co-founder of two technology startup companies in Singapore.
\end{IEEEbiographynophoto}

\end{document}